\documentclass[conference]{IEEEtran}
\usepackage{cite}
\usepackage{amsmath,amssymb,amsfonts}
\usepackage{algorithmic}
\usepackage{graphicx}
\usepackage{textcomp}
\usepackage{xcolor} 
\usepackage{tabularray}
\usepackage{etoolbox}
\usepackage{url}

\newcommand{\redtext}[1]{\textcolor{red}{#1}}
\renewcommand{\redtext}[1]{#1} 

\newcommand{\best}[1]{\textbf{\textcolor{red}{#1}}}

\newcommand{\secondbest}[1]{\textcolor{blue}{\underline{#1}}}

\newlength{\myrowsep}
\makeatletter
\patchcmd{\@makecaption}
  {\scshape}
  {}
  {}
  {}
\makeatletter

\patchcmd{\@makecaption}
  {\\}
  {. \ }
  {}
  {}
\makeatother
\def\tablename{Table}

\def\BibTeX{{\rm B\kern-.05em{\sc i\kern-.025em b}\kern-.08em
    T\kern-.1667em\lower.7ex\hbox{E}\kern-.125emX}}
\begin{document}

\title{TimeDRL: Disentangled Representation Learning for Multivariate Time-Series}

\author{
\IEEEauthorblockN{Ching Chang, Chiao-Tung Chan, Wei-Yao Wang, Wen-Chih Peng, Tien-Fu Chen}
\IEEEauthorblockA{\textit{National Yang Ming Chiao Tung University}\\
Hsinchu, Taiwan \\
blacksnail789521.cs10@nycu.edu.tw, carol.chou.tun.ece05g@g2.nctu.edu.tw, \\
sf1638.cs05@nctu.edu.tw, wcpeng@cs.nycu.edu.tw, tfchen@cs.nycu.edu.tw}
}

\maketitle

\begin{abstract}
Multivariate time-series data in numerous real-world applications (e.g., healthcare and industry) are informative but challenging due to the lack of labels and high dimensionality.
Recent studies in self-supervised learning have shown their potential in learning rich representations without relying on labels, yet they fall short in learning disentangled embeddings and addressing issues of inductive bias (e.g., transformation-invariance).
To tackle these challenges, we propose TimeDRL, a generic multivariate time-series representation learning framework with disentangled dual-level embeddings.
TimeDRL is characterized by three novel features:
(i) disentangled derivation of timestamp-level and instance-level embeddings from patched time-series data using a [CLS] token strategy;
(ii) utilization of timestamp-predictive and instance-contrastive tasks for disentangled representation learning, with the former optimizing timestamp-level embeddings with predictive loss, and the latter optimizing instance-level embeddings with contrastive loss; and
(iii) avoidance of augmentation methods to eliminate inductive biases, such as transformation-invariance from cropping and masking.
Comprehensive experiments on 6 time-series forecasting datasets and 5 time-series classification datasets have shown that TimeDRL consistently surpasses existing representation learning approaches, achieving an average improvement of forecasting by 58.02\% in MSE and classification by 1.48\% in accuracy.
Furthermore, extensive ablation studies confirmed the relative contribution of each component in TimeDRL's architecture, and semi-supervised learning evaluations demonstrated its effectiveness in real-world scenarios, even with limited labeled data.
The code is available at \url{https://github.com/blacksnail789521/TimeDRL}.

\end{abstract}


\section{Introduction}
\label{sec:introduction}
Multivariate time-series data are widely used in various applications, such as forecasting for electric power \cite{informer, electric_forecasting}, activity classification in smartwatches \cite{smartwatch, wisdm}, and anomaly detection in industrial machines \cite{industrial_machine_1, industrial_machine_2}.
These time-series datasets are rich in information, but the patterns within and across temporal dimensions are not discernible by humans, causing a heavy requirement for annotated labels.
Recently, there has been a growing trend among researchers to first learn representations/embeddings from a large amount of unlabeled data using unsupervised representation learning and then to fine-tune these models with a limited amount of labeled data for specific downstream tasks.

Self-supervised learning (SSL) is a prominent method within unsupervised representation learning, which captures generalizable representations from unlabeled data with pretext tasks.
This is beneficial in downstream scenarios with limited labeled records from the generic knowledge of pre-trained models, which has been demonstrated in extracting valuable representations in Natural Language Processing (NLP) \cite{bert,gpt_3} and Computer Vision (CV) \cite{simclr,byol,simsiam}.
However, applying SSL directly in the time-series domain faces two challenges.


\begin{figure}[t]
    \centering
    \includegraphics[width=\columnwidth]{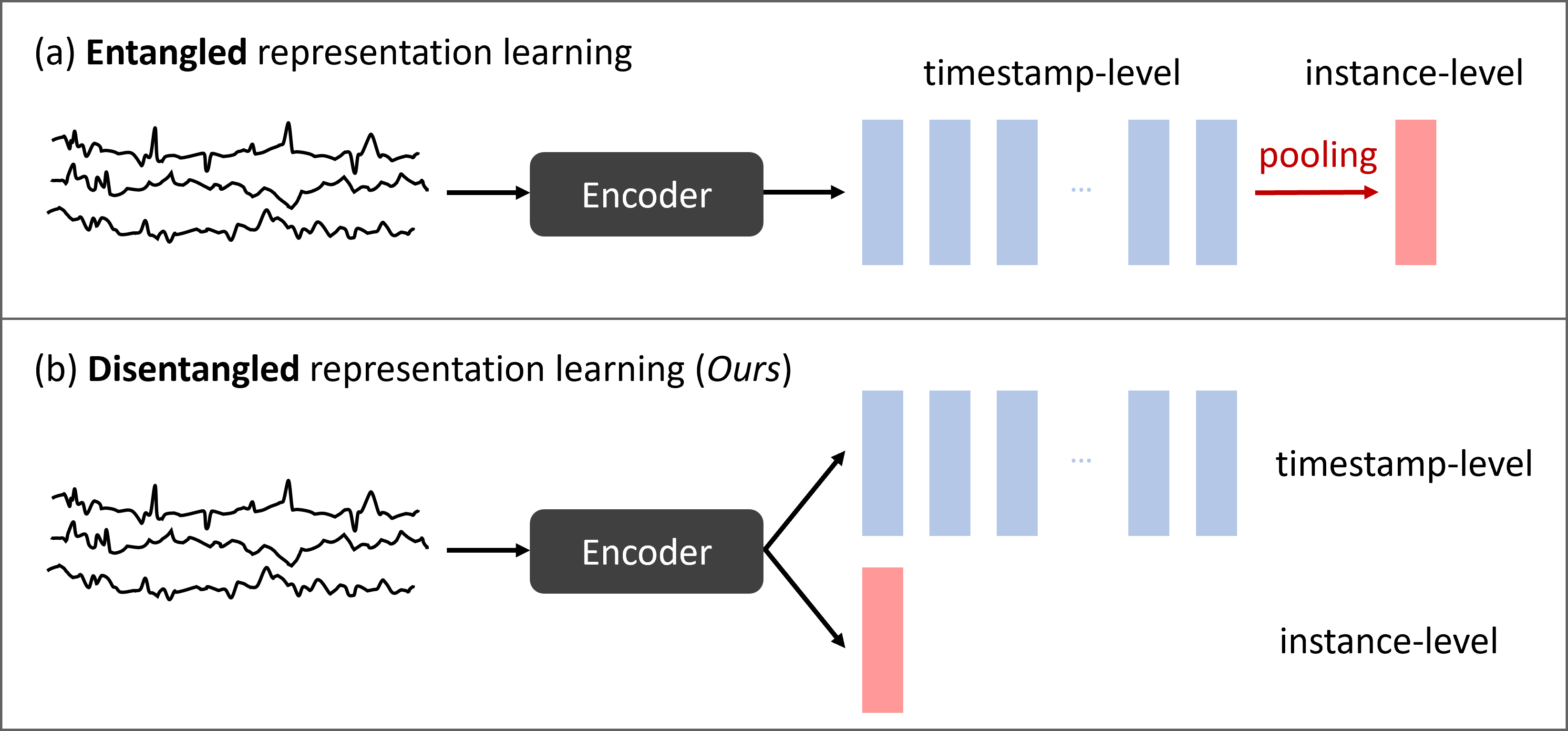}
    \caption{\textbf{Two categories of representation learning in the time-series domain.} The top section (a) represents entangled representation learning, where timestamp-level embeddings are first derived and then a pooling method is applied to extract instance-level embeddings. The bottom section (b) illustrates disentangled representation learning, which involves deriving timestamp-level and instance-level embeddings in a disentangled manner. Our proposed TimeDRL employs disentangled derivation of timestamp-level and instance-level embeddings.}
    \label{fig:DRL}
\end{figure}

\begin{figure*}
    \centering
    \includegraphics[width=2.0\columnwidth]{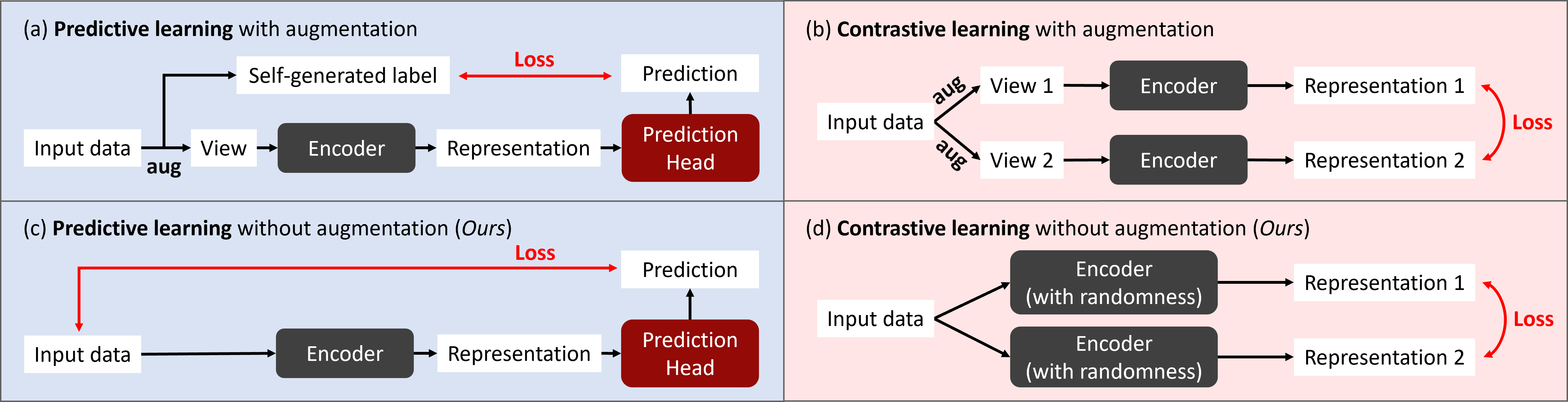}
    \caption{\textbf{Two categories of self-supervised learning.} The left sections (a) and (c) represent predictive learning, utilizing a single representation to predict inherent data characteristics. The right sections (b) and (d) illustrate contrastive learning, emphasizing the distinction of subtle differences between data samples. Our proposed TimeDRL avoids using any augmentation methods in both categories to enable robust learning and prevent inductive bias ((c) and (d)).}
    \label{fig:ssl_two_categories}
\end{figure*}

The first challenge in SSL for time-series data is learning \textit{disentangled dual-level representations}.
Existing approaches focus on deriving either timestamp-level \cite{ts2vec, simts} or instance-level embeddings \cite{tnc, t_loss, mhccl}, but not both at the same time.
However, these two types of embeddings serve distinct purposes: timestamp-level embeddings are effective for forecasting and anomaly detection, whereas instance-level embeddings are suited for classification and clustering tasks \cite{ssl_time_series}.
Although we can theoretically avoid explicitly deriving instance-level embeddings by extracting them from timestamp-level embeddings using pooling methods  (as illustrated in Fig. \ref{fig:DRL}(a)) \cite{ts2vec}, this approach often results in the anisotropy problem \cite{anisotropy_1, anisotropy_2, anisotropy_3}, where the embeddings are confined to a narrow cone region in the embedding space, thus limiting their expressiveness.
To the best of our knowledge, how to disentangle instance-level embeddings from timestamp-level embeddings in the time-series domain remains an unexplored problem.

The second challenge lies in the \textit{inductive bias}.
Inductive bias refers to the assumptions and prior knowledge that data augmentation methods are adopted in the learning process \cite{ts2vec} to enhance a model's ability to generalize to unseen data.
Directly applying data augmentation methods from other domains (e.g., image colorization \cite{image_color} and rotation \cite{image_rotation} in CV, and masking \cite{bert} and synonym replacement \cite{synonym_replacement} in NLP) to the time-series domain is not practical.
These methods introduce inductive biases that are not suitable for time-series data.
For instance, rotation is a common augmentation technique for images.
However, it's unsuitable for time-series data, as it can disrupt chronological order and essential temporal patterns, leading to distorted interpretations.
While there are time-series-specific augmentations like scaling, permutation \cite{ts_tcc}, and cropping \cite{ts2vec}, these techniques are based on the assumption that models benefit from learning transformation-invariant representations.
However, this assumption can be limiting, as it overlooks the unique and diverse characteristics of different time-series datasets. 

To address the above issues, we propose TimeDRL, a generic multivariate \textbf{Time}-series framework with \textbf{D}isentangled \textbf{R}epresentation \textbf{L}earning for dual-level embeddings.
By disentangling timestamp-level and instance-level embeddings  (as illustrated in Fig. \ref{fig:DRL}(b)), TimeDRL is applicable across various time-series downstream tasks.
This method involves aggregating adjacent time steps to form patched time-series data, followed by appending a [CLS] token at the beginning to represent the instance-level representation.
Notably, using \textit{patched} time-series data, as opposed to \textit{point-level}, allows us to extend the horizon without significantly increasing training costs, ensuring that the [CLS] token captures more comprehensive semantic information.
To optimize the encoder network, TimeDRL employs two pretext tasks: a timestamp-predictive task for optimizing timestamp-level embeddings, and an instance-contrastive task for optimizing instance-level embeddings.
\redtext{
Specifically, two pretext tasks are designed in a disentangled manner, each of which is specifically tailored to optimize its respective embedding level, ensuring that the optimization of one embedding type does not interfere with the other.
}
To mitigate the inductive bias of transformation-invariance, we refrain from directly applying any augmentation to the data in both predictive and contrastive learning tasks, as illustrated in (c) and (d) of Fig. \ref{fig:ssl_two_categories}.
In the timestamp-predictive task, we opt to use reconstruction error on patched time-series data, notably without masking any input data.
For the instance-contrastive task, we use dropout layers to introduce variation. 
By leveraging the inherent randomness of dropout layers, we can generate distinct views of the embeddings from the same input without relying on any external augmentation methods.

In summary, the paper’s main contributions are as follows:

\begin{itemize}
\item \textbf{Universal Applicability with Disentangled Dual-Level Embeddings}: We introduce TimeDRL, a multivariate time-series representation learning framework that operates on disentangled dual-level embeddings. This design enables broad applicability across a variety of time-series downstream tasks. TimeDRL leverages a dedicated [CLS] token in conjunction with patched time-series data, allowing the instance-level embedding to capture a more comprehensive range of semantic information by extending the time-series horizon.
\item \textbf{Two Pretext Tasks}: TimeDRL employs a timestamp-predictive task to optimize timestamp-level embeddings and an instance-contrastive task to optimize instance-level embeddings, ensuring effective learning on both levels.
\item \textbf{Mitigation of Inductive Bias}: To avoid inductive bias, TimeDRL utilizes a reconstruction error approach without masking in the timestamp-predictive task, and leverages the randomness of dropout layers in the instance-contrastive task, ensuring unbiased data representation.
\item \textbf{Effective Performance Across 11 Real-World Benchmarks}: TimeDRL is consistently superior to state-of-the-art performance in both time-series forecasting and classification benchmarks, demonstrating the generalizability of the proposed approach in real-world applications.
\end{itemize}

\section{Related Work}

\subsection{Foundational Concepts of Self-Supervised Learning: A Pre-Time-Series Perspective}
SSL methods have demonstrated the effectiveness of learning generic representations from designing pretext tasks, which can be divided into two categories: \textit{predictive} and \textit{contrastive} learning \cite{ssl_graph_survey}. 
As illustrated in Fig. \ref{fig:ssl_two_categories}, predictive learning utilizes a single representation to predict characteristics inherent within the data.
In contrast, contrastive learning emphasizes distinguishing subtle differences between data samples by calculating loss on pairs of representations.
Siamese networks \cite{siamese} are weight-sharing neural networks, and are commonly used in contrastive learning to process pairs of inputs simultaneously.

The concepts of prediction and contrastive learning first emerged in the fields of NLP and CV.
In NLP, BERT \cite{bert} introduces predictive tasks with masked language modeling and next sentence prediction to learn semantically rich representations, while GPT \cite{gpt_3} uses an autoregressive predictive task to exhibit its few-shot learning capabilities.
SimCSE \cite{simcse} applies contrastive tasks to improve sentence-level embeddings from the [CLS] token. It utilizes dropout layers to introduce variation, without relying on any external augmentation methods.
In CV, SimCLR \cite{simclr} uses contrastive learning to learn detailed representations by treating augmented views of the same instance as positive pairs and all other instances within the minibatch as negative pairs.
BYOL \cite{byol} and SimSiam \cite{simsiam}, on the other hand, facilitate the additional prediction head with the stop-gradient strategy to avoid negative samples and eliminate the need for large batch sizes.

In recent years, SSL technology has expanded into new domains like tabular data and Graph Neural Networks (GNNs).
In the tabular data domain, VIME \cite{vime} introduces a predictive task involving mask vector estimation with an autoencoder architecture, and SCARF \cite{scarf} employs contrastive learning for detailed representation learning.
In GNNs, GraphCL \cite{graphcl} and BGRL \cite{bgrl} have adopted contrastive learning to GNNs and have shown promising results in tasks such as graph classification and edge prediction.
However, adopting SSL techniques across domains often brings inductive bias.
Data augmentation methods like image colorization \cite{image_color} and rotation \cite{image_rotation} in CV, or masking \cite{bert} and synonym replacement \cite{synonym_replacement} in NLP, may introduce biases unsuitable for the target domain.
To address this, various works have proposed domain-specific solutions.
For instance, MTR \cite{mtr} in tabular data proposes an augmentation method tailored for tabular formats.
In GNNs, SimGRACE \cite{simgrace} completely avoids the use of data augmentation.
Following this insight, TimeDRL avoids data augmentation methods across all pretext tasks to eliminate any potential inductive biases.

\subsection{Self-Supervised Learning for Time-Series Data}
Self-supervised learning for the representation of time-series data has been gaining traction in recent years.
T-Loss \cite{t_loss} uses a triplet loss with time-based negative sampling to learn representations of time-series data.
TNC \cite{tnc} uses the Augmented Dickey-Fuller (ADF) statistical test to determine temporal neighborhoods and uses Positive-Unlabeled (PU) learning to reduce the impact of sampling bias.
TS-TCC \cite{ts_tcc} first creates two views through strong and weak augmentations, then learns representations through cross-view temporal and contextual contrasting. 
TS2Vec \cite{ts2vec} concentrates on distinguishing multi-scale contextual information at both instance and timestamp levels, establishing itself as the first universal framework effective across various time-series tasks.
TF-C \cite{tf_c} proposes a method of encoding a temporal-based neighborhood close to its frequency-based neighborhood through time-frequency consistency.
MHCCL \cite{mhccl} leverages semantic information from a hierarchical structure of multiple latent partitions in multivariate time-series, enhanced by hierarchical clustering for more informative positive and negative pairing.
SimTS \cite{simts} offers a simplified approach for enhancing time-series forecasting by learning to predict the future from the past in a latent space without reliance on negative pairs or specific assumptions about the time-series characteristics. 
Several efforts to adopt self-supervised learning (SSL) for time-series data have primarily focused on deriving instance-level embeddings by extracting them from timestamp-level embeddings using pooling methods \cite{ts2vec}.
However, this approach often leads to an anisotropy problem, where embeddings are constrained to a narrow region in the embedding space, limiting their expressiveness \cite{anisotropy_1, anisotropy_2, anisotropy_3}.
TimeDRL addresses this issue by disentangling timestamp-level and instance-level embeddings.

\begin{figure*}
    \centering
    \includegraphics[width=\linewidth]{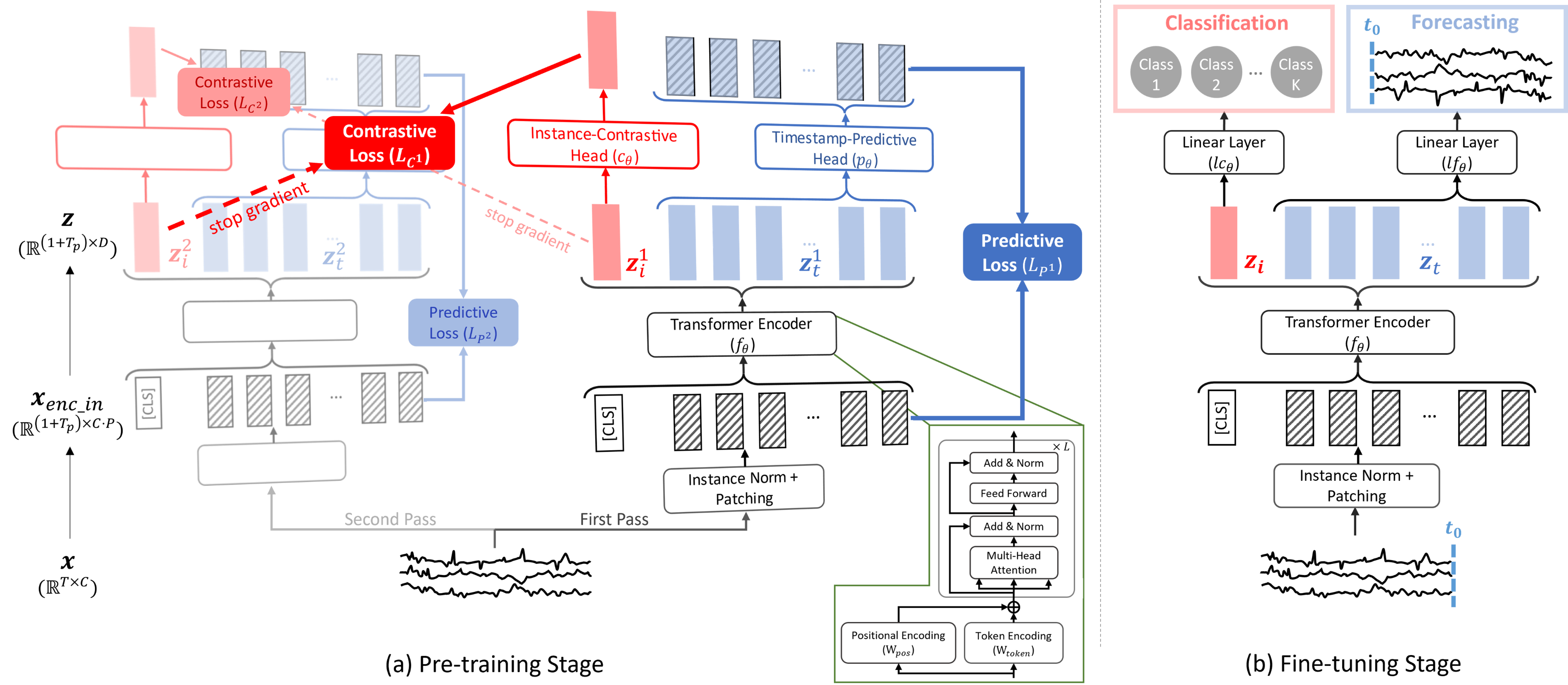}
    \caption{\redtext{\textbf{TimeDRL framework}. The framework is illustrated in two sections: (a) \textit{Pre-training Stage} and (b) \textit{Fine-tuning Stage}. In the pre-training stage (a), we adopt a Siamese network architecture with a Transformer encoder as our encoder \(f_\theta\) to generate two distinct views of embeddings from the same input. This is achieved by exploiting the inherent randomness of the dropout layers within the encoder, thus eliminating the need for data augmentations. The fine-tuning stage (b) demonstrates the application of these embeddings in downstream tasks, showcasing the adaptability of TimeDRL's pre-trained embeddings for time-series forecasting and classification.}}
    \label{fig:framework}
\end{figure*}

\section{Problem Formulation}
Given an \textit{unlabeled} set of \( N \) multivariate time-series samples \( D_u = \{ \textbf{x}^{(n)} \}_{n=1}^{N} \), the objective is to develop an encoder network \( f_\theta \) capable of mapping each sample \( \textbf{x}^{(n)} \) to its corresponding representation \( \textbf{z}^{(n)} \).
For simplicity, the superscript \( (n) \) denoting the sample index is omitted in subsequent descriptions.
The encoder network \( f_\theta \) is designed to produce either timestamp-level or instance-level embeddings based on the task requirement:
\begin{enumerate}
    \item \textbf{Timestamp-Level Embedding}: Each input time-series sample \( \textbf{x} \in \mathbb{R}^{T \times C} \) is encoded into \( \textbf{z}_t \in \mathbb{R}^{T \times D_t} \), where \( T \) represents the sequence length, \( C \) represents the number of features, and \( D_t \) represents the dimension of the timestamp-level embedding. Each timestamp of \( \textbf{z}_t \) represents the specific characteristics and information corresponding to each timestamp of \( \textbf{x} \), enabling a detailed and granular understanding of the time-series data at each individual timestamp.
    \item \textbf{Instance-Level Embedding}: Each input sample \( \textbf{x} \in \mathbb{R}^{T \times C} \) is encoded into an instance-level embedding \( \textbf{z}_i \in \mathbb{R}^{D_i} \), with \( D_i \) being the dimension of the instance-level embedding. Here, \( T \) and \( C \) retain their meanings as sequence length and number of features, respectively. The embedding \( \textbf{z}_i \) represents the overall information of the entire series \( \textbf{x} \), enabling a comprehensive view of the data.
\end{enumerate}

\section{The Proposed TimeDRL}
Fig. \ref{fig:framework} presents the overall architecture of the TimeDRL framework.
We detail the methodology for obtaining both timestamp-level embeddings \(\textbf{z}_t\) and instance-level embeddings \(\textbf{z}_i\) from an input time-series sample \(\textbf{x}\), employing a Transformer encoder \(f_\theta\) as our backbone encoder (Section \ref{sec:dual_level_embeddings}).
Later, we introduce two essential pretext tasks.
We first explore our strategy for optimizing timestamp-level embeddings \(\textbf{z}_t\) using a timestamp-predictive task, specifically without implementing any data augmentations (Section \ref{sec:predictive_learning}).
Subsequently, our approach for optimizing instance-level embeddings \(\textbf{z}_i\) through an instance-contrastive task is discussed, highlighting the avoidance of augmentations and negative sample pairs (Section \ref{sec:contrastive_learning}).

\subsection{Disentangled Dual-Level Embeddings}
\label{sec:dual_level_embeddings}
Transformers have achieved notable success in downstream time-series tasks \cite{fedformer, patchtst, anomaly_transformer, gtn}, but in the realm of self-supervised learning for time-series, CNN-based \cite{ts_tcc, ts2vec} and RNN-based \cite{tnc} models are generally preferred over Transformers.
This indicates that the full potential of Transformers in time-series representation learning has yet to be fully exploited.
Moreover, models such as BERT \cite{bert} and RoBERTa \cite{roberta} have proven successful in generating high-quality sentence embeddings, making the Transformer encoder a suitable choice for our framework.
This decision is further validated by the similarity between extracting sentence-level embeddings from token-level embeddings in BERT and deriving instance-level embeddings from timestamp-level embeddings in time-series data.
Given these considerations, the Transformer encoder has been adopted as the primary architecture for TimeDRL, as illustrated in the bottom right part of Fig. \ref{fig:framework}.
Furthermore, we adopt the \textit{patching} concept from PatchTST \cite{patchtst}, which aggregates adjacent time steps into a single patch-based token.
This technique substantially cuts down the context window size needed for Transformers, significantly reducing training costs and greatly enhancing the stability of the training process.

In BERT and RoBERTa, the [CLS] token is employed to capture sentence-level embeddings, which motivated us to adopt it to extract instance-level embeddings in the time-series domain.
While it is theoretically possible to derive instance-level embeddings from timestamp-level embeddings through pooling methods (e.g., global average pooling) \cite{ts2vec}, this approach can lead to the anisotropy problem \cite{anisotropy_1, anisotropy_2, anisotropy_3}.
This issue arises when the timestamp-level embeddings are restricted to a narrow region in the embedding space, reducing their capacity to effectively capture diverse information.
Studies in NLP \cite{simcse, diffcse} have indicated that compared to traditional pooling methods, optimizing the [CLS] token through contrastive learning can yield better results.
This finding aligns with our experimental outcomes, as detailed in Table \ref{tab:ablation_pooling}.

Given an input time-series sample \( \textbf{x} \in \mathbb{R}^{T \times C} \), we first apply instance normalization (\(\text{IN}\)) \cite{revin} and patching \cite{patchtst} to produce a series of patches \( \textbf{x}_{patched} \in \mathbb{R}^{T_p \times C \cdot P} \):
\begin{equation}
    \textbf{x}_{patched} = \text{patching}(\text{IN(\textbf{x})}).
\end{equation}
The patching process reduces the input sample's time dimension from \(T\) to \(T_p\), where \(T_p\) denotes the number of patches, and concurrently expands the feature dimension from \(C\) to \(C \cdot P\), with \(P\) representing the patch length.
Later, a [CLS] token \( \in \mathbb{R}^{C \cdot P} \) is appended at the beginning of these patches, resulting in the final input \(\textbf{x}_{enc\_in} \in \mathbb{R}^{(1 + T_p) \times C \cdot P}\) for the encoder \(f_\theta\):
\begin{equation}
    \textbf{x}_{enc\_in} = \text{concatenate}(\text{[CLS]}, \textbf{x}_{patched}).
\end{equation}
After receiving the encoder input \(\textbf{x}_{enc\_in}\), we pass it into the encoder \(f_\theta\), which comprises a linear token encoding layer \( W_{token} \in \mathbb{R}^{D \times C \cdot P} \), a learnable linear positional encoding layer \( PE \in \mathbb{R}^{(1 + T_p) \times D} \), and a series of Transformer blocks \(\text{TBs}\) (with \(L\) blocks in total).
This process yields the final embeddings \( \textbf{z} \in \mathbb{R}^{(1 + T_p) \times D} \):
\begin{equation}
    \textbf{z} = \text{TBs}(\textbf{x}_{enc\_in}W_{token}^\top + PE).
\end{equation}
It is important to note that within the Transformer encoder \(f_\theta\) (including its encoding layers), the dimensions of its input \(\textbf{x}_{enc\_in}\) and output \(\textbf{z}\) differ only in their feature dimension, transitioning from \(C \cdot P\) to \(D\), where \(D\) represents the dimension of the Transformer's latent space.

After obtaining the time-series embedding \( \textbf{z} \) from the input \(\textbf{x}\) using the encoder \(f_\theta\), our next step is to extract the timestamp-level embedding \( \textbf{z}_t \in \mathbb{R}^{T_p \times D} \) and the instance-level embedding \( \textbf{z}_i \in \mathbb{R}^{D} \).
The extraction process is straightforward: the embedding corresponding to the first token, which is the [CLS] token, is designated as the instance-level embedding \(\textbf{z}_i\), while the subsequent embeddings are considered as the timestamp-level embedding \(\textbf{z}_t\):
\begin{equation}
\textbf{z}_i = \textbf{z}[0, :],
\end{equation}
\begin{equation}
\textbf{z}_t = \textbf{z}[1:T_p+1, :].
\end{equation}

\subsection{Timestamp-Predictive Task in TimeDRL}
\label{sec:predictive_learning}
To capture the relations between timestamps, we develop a timestamp-predictive task to derive the timestamp-level embeddings through predictive loss.
As discussed in Section \ref{sec:introduction}, the aim of the timestamp-predictive task is to derive timestamp-level embedding without introducing inductive bias.
Most existing approaches in NLP \cite{bert,roberta} or in time-series domain \cite{patchtst} utilize masked language modeling (MLM) to learn the semantics of the token embeddings, where masking is an augmentation strategy that causes inductive bias.
The underlying assumption of data augmentation is that the encoder should maintain consistent embeddings despite input transformations.
However, given the vast diversity in characteristics among different time-series datasets, we design TimeDRL as a universal framework that avoids any consideration of transformation-invariance.
Therefore, we introduce the non-augmented timestamp-predictive task in TimeDRL by focusing on the reconstruction of the patched time-series data without augmentation methods.

Given a timestamp-level embedding \( \textbf{z}_t \), it is first processed through a timestamp-predictive head \( p_\theta \) (a linear layer without an activation function) to generate a prediction. 
To ensure that this prediction accurately reconstructs the original patched data \( \textbf{x}_{patched} \), we use the Mean Squared Error (MSE) as the loss function.
The predictive loss \(\mathcal{L}_{P}\) is calculated as the MSE between \( \textbf{x}_{patched} \) and the predicted output:
\begin{equation}
\mathcal{L}_{P} = \text{MSE} (\textbf{x}_{patched}, p_\theta(\textbf{z}_t)).
\end{equation}
It is worth noting that the instance-level embeddings \( \textbf{z}_i \) are not updated from the MSE loss.

While we initially discuss predictive loss \(\mathcal{L}_{P}\) for a single timestamp-level embedding \(\textbf{z}_t\), it is important to note that our framework generates two views of representations \(\textbf{z}^1\) and \(\textbf{z}^2\), as the input data \(\textbf{x}\) is processed twice through the encoder \(f_\theta\) (discussed in Section \ref{sec:contrastive_learning}).
This enables us to apply predictive learning to both timestamp-level embeddings \(\textbf{z}^1_t\) and \(\textbf{z}^2_t\).
Thus, the predictive loss for each representation is calculated as follows:
\begin{equation}
\mathcal{L}_{P^1} = MSE (\textbf{x}_{patched}, p_\theta(\textbf{z}^1_t)),
\end{equation}
\begin{equation}
\mathcal{L}_{P^2} = MSE (\textbf{x}_{patched}, p_\theta(\textbf{z}^2_t)),
\end{equation}
and the total predictive loss \(\mathcal{L}_P\) will be the average of \(\mathcal{L}_{P^1}\) and \(\mathcal{L}_{P^2}\):
\begin{equation}
\mathcal{L}_P = \frac{1}{2}\mathcal{L}_{P^1} + \frac{1}{2}\mathcal{L}_{P^2}.
\end{equation}

\subsection{Instance-Contrastive Task in TimeDRL}
\label{sec:contrastive_learning}

To capture the overall information of the entire series, we develop an instance-contrastive task to derive the instance-level embeddings through contrastive loss.
In contrastive learning, two different views of embeddings are required to compute the loss.
To align with our commitment to avoid data augmentations, we use dropout layers within the backbone encoder to introduce randomness on the output embedding \cite{simcse}.
By passing the data through the encoder twice, two distinct views of the embeddings are generated from the same input:
\begin{equation}
    \text{First pass:} \quad \textbf{z}^1 = f_\theta(\textbf{x}_{patched}),
\end{equation}
\begin{equation}
    \text{Second pass:} \quad \textbf{z}^2 = f_\theta(\textbf{x}_{patched}).
\end{equation}
Then, for each embedding, we extract the first position as our instance-level embedding.
\begin{equation}
\textbf{z}^1_i = \textbf{z}^1[0, :],
\end{equation}
\begin{equation}
\textbf{z}^2_i = \textbf{z}^2[0, :],
\end{equation}
This strategy ensures that no external data augmentation methods are used, thereby avoiding introducing inductive bias.

In addition, naively using this technique fails to address \textit{sampling bias} in contrastive learning.
Sampling bias occurs when randomly selected negative samples are similar to the positive samples, which is a common scenario in the time-series domain due to the presence of periodic patterns.
\redtext{
To that end, we focus exclusively on positive samples and remove the use of negative samples to address the challenge of sampling bias.
The proposed method incorporates an additional prediction head and integrates a stop-gradient operation to prevent model collapse \cite{simsiam, avoid_collapse}.
}
Furthermore, this negative-free approach in contrastive learning removes the need for large batch sizes, which are usually essential to gather enough negative samples for stable training.

After obtaining two instance-level embeddings \( \textbf{z}^1_i \) and \( \textbf{z}^2_i \) from the same input via the randomness in dropout layers, each embedding is processed through an instance-contrastive head \( c_\theta \) (a two-layer bottleneck MLP with BatchNorm and ReLU in the middle) to produce \( \hat{\textbf{z}}^1_i \) and \( \hat{\textbf{z}}^2_i \):
\begin{equation}
\hat{\textbf{z}}^1_i = c_\theta(\textbf{z}^1_i),
\end{equation}
\begin{equation}
\hat{\textbf{z}}^2_i = c_\theta(\textbf{z}^2_i).
\end{equation}
When calculating the contrastive loss, our objective is to align \(\hat{\textbf{z}}^1_i\) with \(\textbf{z}^2_i\).
This treatment of \(\textbf{z}^2_i\) as a constant in the contrastive loss ensures that the model updates are based solely on the predicted \(\hat{\textbf{z}}^1_i\), without receiving gradients from \(\textbf{z}^2_i\).
The loss is computed using negative cosine similarity as:
\begin{equation}
\mathcal{L}_{C^1} = -\text{cosine}(\hat{\textbf{z}}^1_i, \text{stop\_gradient}(\textbf{z}^2_i)),
\end{equation}
where \( \text{stop\_gradient} \) represents the stop-gradient operation.
Similarly, to symmetrically optimize the network, we also calculate the loss between \(\hat{\textbf{z}}^2_i\) and \(\textbf{z}^1_i\), applying the stop-gradient operation to \(\textbf{z}^1_i\):
\begin{equation}
\mathcal{L}_{C^2} = -\text{cosine}(\hat{\textbf{z}}^2_i, \text{stop\_gradient}(\textbf{z}^1_i)).
\end{equation}
The total contrastive loss \(\mathcal{L}_C\) is the average of \(\mathcal{L}_{C^1}\) and \(\mathcal{L}_{C^2}\):
\begin{equation}
\mathcal{L}_C = \frac{1}{2}\mathcal{L}_{C^1} + \frac{1}{2}\mathcal{L}_{C^2}.
\end{equation}

Finally, the timestamp-predictive task and the instance-contrastive task are jointly trained, and \(\lambda\) is used to adjust between the two losses:
\begin{equation}
\label{eq:overall_loss}
\mathcal{L} = \mathcal{L}_{P} + \lambda \cdot \mathcal{L}_{C}.
\end{equation}

\section{Experiments}
To ensure reproducibility, we utilize time-series datasets from reputable sources for our evaluations.
Our TimeDRL framework is evaluated in two key areas: \textit{forecasting}, which tests the effectiveness of timestamp-level embeddings, and \textit{classification}, focusing on the utility of instance-level embeddings.
For time-series forecasting, we benchmark TimeDRL against 6 baseline models across 6 different datasets.
Similarly, in the time-series classification domain, the model is compared with 6 baselines on 5 datasets.

We start with a linear evaluation to assess the effectiveness of both timestamp-level and instance-level embeddings learned by TimeDRL.
The results indicate that training a linear layer on these embeddings surpasses previous state-of-the-art methods.
Then, we showcase the model's performance in semi-supervised learning scenarios, highlighting its efficacy when dealing with limited labeled data alongside a substantial volume of unlabeled data.
Lastly, a series of ablation studies are conducted to highlight the significance of each component within the TimeDRL framework.

\begin{table}[t]
\centering
\caption{\textbf{Statistical overview of the 7 datasets for time-series forecasting.}}
\label{tab:forecasting_dataset}
\begin{tblr}{
  rowsep = 1.5pt,
  cells={font=\small},
  cells = {c},
  vline{2} = {-}{},
  hline{1,2,6} = {-}{},
}
Datasets & Features & Timesteps & Frequency\\
ETTh1 \& ETTh2 & 7 & 17,420 & 1 hour\\
ETTm1 \& ETTm2 & 7 & 69,680 & 5 min\\
Exchange & 8 & 7,588 & 1 day\\
Weather & 21 & 52,696 & 10 min
\end{tblr}
\end{table}

\begin{table}[t]
\centering
\caption{\textbf{Statistical overview of the 5 datasets for time-series classification.}}
\label{tab:classification_dataset}
\begin{tblr}{
  rowsep = 1.5pt,
  cells={font=\small},
  cells = {c},
  vline{2} = {-}{},
  hline{1,2,7} = {-}{},
}
Datasets & Samples & Features & Classes & Length\\
FingerMovements & 416 & 28 & 2 & 50\\
PenDigits & 10,992 & 2 & 10 & 8\\
HAR & 10,299 & 9 & 6 & 128\\
Epilepsy & 11,500 & 1 & 2 & 178\\
WISDM & 4,091 & 3 & 6 & 256
\end{tblr}
\end{table}

\subsubsection{\textbf{Datasets}}
\paragraph{Datasets for Time-Series Forecasting}
For our time-series forecasting analysis, we conducted experiments on 6 real-world, publicly available benchmark datasets.
Detailed in Table \ref{tab:forecasting_dataset} is an overview of the characteristics of each dataset used in the forecasting experiments.
This includes the number of features, the total dataset length, and the sampling frequency.

\textbf{ETT} \cite{informer} captures long-term electric power deployment data.
These datasets include two hourly-sampled datasets (ETTh1, ETTh2) and two 15-minute-sampled datasets (ETTm1, ETTm2), spanning over two years from different provinces in China.
The ETT datasets comprise one oil temperature feature along with 6 power load features.
All features are utilized for multivariate forecasting, whereas only the oil temperature feature is employed for univariate forecasting.
\textbf{Exchange} \cite{exchange_rate} comprises the daily exchange rates of 8 foreign countries, spanning from 1990 to 2016.
These countries include Australia, Britain, Canada, Switzerland, China, Japan, New Zealand, and Singapore.
For multivariate forecasting, we use the data from all these countries, whereas for univariate forecasting, we specifically focus on Singapore.
\textbf{Weather}\footnote{https://www.ncei.noaa.gov/data/local-climatological-data/} provides local climatological data for nearly 1,600 U.S. areas over 4 years.
Each record includes 11 weather variables along with the target feature 'web bulb.'
For multivariate forecasting, all features are considered, whereas for univariate forecasting, we focus specifically on the 'web bulb' feature.

\paragraph{Datasets for Time-Series Classification}
For our time-series classification analysis, we performed experiments on 5 real-world, publicly accessible benchmark datasets.
Table \ref{tab:forecasting_dataset} provides an overview of each dataset's characteristics, including the number of time-series samples, the number of features, the number of classes, and the length of each sample.

\textbf{HAR} \cite{har} comprises sensor data from 30 subjects performing 6 activities.
The data were gathered using a Samsung Galaxy S2 device, and the objective was to predict the activity based on accelerometer and gyroscope measurements.
\textbf{WISDM} \cite{wisdm} includes time-series data from accelerometers and gyroscopes in smartphones and smartwatches.
It was collected while 51 test subjects performed 18 different activities, each for a duration of 3 minutes.
\textbf{Epilepsy} \cite{epilepsy} contains EEG recordings from 500 individuals, recorded using a single-channel EEG sensor at a frequency of 174 Hz.
The data include 23.6 seconds of brain activity recordings for each subject, classified as either having epilepsy or not.
\textbf{PenDigits} \cite{pendigits} addresses a handwritten digit classification task where 44 writers drew the digits 0 to 9 and the x and y coordinates were recorded.
The data were recorded at 500x500 pixel resolution and then resampled to 8 spatial points.
\textbf{FingerMovements} \cite{fingermovements} is a dataset recorded from a subject performing self-paced key typing on a computer keyboard.
The task consisted of three 6-minute sessions conducted on the same day with breaks in between, with typing performed at an average speed of one key per second.

\subsubsection{\textbf{Evaluation Metrics}}
\paragraph{Evaluation Metrics for Time-Series Forecasting}
In time-series forecasting, we primarily use Mean Squared Error (MSE) and Mean Absolute Error (MAE) as our evaluation metrics. 
The Mean Squared Error (MSE) is defined as:
\begin{equation}
\text{MSE} = \frac{1}{N} \sum_{n=1}^{N} (\textbf{y}^{(n)} - \hat{\textbf{y}}^{(n)})^2.
\end{equation}
where \( \textbf{y}^{(n)} \) represents the actual future sequence value corresponding to the input \( \textbf{x}^{(n)} \), \( \hat{\textbf{y}}^{(n)} \) is the predicted value for the same input, and \( N \) denotes the total number of samples.
The Mean Absolute Error (MAE) is defined as:
\begin{equation}
\text{MAE} = \frac{1}{N} \sum_{n=1}^{N} |\textbf{y}^{(n)} - \hat{\textbf{y}}^{(n)}|.
\end{equation}

\begin{table*}[ht]
\centering
\caption{\redtext{\textbf{Linear evaluation on multivariate time-series forecasting.} We use prediction lengths \(T \in \{24, 48, 168, 336, 720\}\) for ETTh1, ETTh2, Exchange, and Weather; and \(T \in \{24, 48, 96, 228, 672\}\) for ETTm1 and ETTm2. The best results are in \best{bold}, while the second-best are \secondbest{underlined}.}}
\label{tab:linear_eval_forecasting_multivariate}
\begin{tblr}{
  rowsep = 1.7pt,
  colsep = 5pt,
  cells = {c},
  cells={font=\small},
  cell{1}{1} = {c=2,r=2}{},
  cell{1}{3} = {c=10}{},
  cell{1}{13} = {c=4}{},
  cell{2}{3} = {c=2}{},
  cell{2}{5} = {c=2}{},
  cell{2}{7} = {c=2}{},
  cell{2}{9} = {c=2}{},
  cell{2}{11} = {c=2}{},
  cell{2}{13} = {c=2}{},
  cell{2}{15} = {c=2}{},
  cell{3}{1} = {c=2}{},
  cell{34}{1} = {c=2}{},
  cell{4}{1} = {r=5}{},
  cell{9}{1} = {r=5}{},
  cell{14}{1} = {r=5}{},
  cell{19}{1} = {r=5}{},
  cell{24}{1} = {r=5}{},
  cell{29}{1} = {r=5}{},
  vline{2-3,5,7,9,11,13,15} = {1-39}{},
  hline{1-4,9,14,19,24,29,34} = {-}{},
}
Methods &  & Unsupervised
  Representation Learning &  &  &  &  &  &  &  &  &  & End-to-end
  Forecasting &  &  & \\
 &  & TimeDRL &  & SimTS &  & TS2Vec &  & TNC &  & CoST &  & Informer &  & TCN & \\
Metric &  & MSE & MAE & MSE & MAE & MSE & MAE & MSE & MAE & MSE & MAE & MSE & MAE & MSE & MAE\\
ETTh1 & 24 & \best{0.327} & \best{0.378} & \secondbest{0.377} & \secondbest{0.422} & 0.59 & 0.531 & 0.708 & 0.592 & 0.386 & 0.429 & 0.577 & 0.549 & 0.583 & 0.547\\
 & 48 & \best{0.353} & \best{0.392} & \secondbest{0.427} & \secondbest{0.454} & 0.624 & 0.555 & 0.749 & 0.619 & 0.437 & 0.464 & 0.685 & 0.625 & 0.670 & 0.606\\
 & 168 & \best{0.418} & \best{0.427} & \secondbest{0.638} & \secondbest{0.577} & 0.762 & 0.639 & 0.884 & 0.699 & 0.643 & 0.582 & 0.931 & 0.752 & 0.811 & 0.680\\
 & 336 & \best{0.437} & \best{0.446} & 0.815 & 0.685 & 0.931 & 0.728 & 1.020 & 0.768 & \secondbest{0.812} & \secondbest{0.679} & 1.128 & 0.873 & 1.132 & 0.815\\
 & 720 & \best{0.446} & \best{0.461} & \secondbest{0.956} & \secondbest{0.771} & 1.063 & 0.799 & 1.157 & 0.830 & 0.970 & \secondbest{0.771} & 1.215 & 1.869 & 1.165 & 0.813\\
ETTh2 & 24 & \best{0.183} & \best{0.279} & \secondbest{0.336} & \secondbest{0.434} & 0.424 & 0.489 & 0.612 & 0.595 & 0.447 & 0.502 & 0.720 & 0.665 & 0.935 & 0.754\\
 & 48 & \best{0.229} & \best{0.308} & \secondbest{0.564} & \secondbest{0.571} & 0.619 & 0.605 & 0.840 & 0.716 & 0.699 & 0.637 & 1.457 & 1.001 & 1.300 & 0.911\\
 & 168 & \best{0.334} & \best{0.376} & \secondbest{1.407} & \secondbest{0.926} & 1.845 & 1.074 & 2.359 & 1.213 & 1.549 & 0.982 & 3.489 & 1.515 & 4.017 & 1.579\\
 & 336 & \best{0.372} & \best{0.407} & \secondbest{1.640} & \secondbest{0.996} & 2.194 & 1.197 & 2.782 & 1.349 & 1.749 & 1.042 & 2.723 & 1.340 & 3.460 & 1.456\\
 & 720 & \best{0.395} & \best{0.428} & \secondbest{1.878} & \secondbest{1.065} & 2.636 & 1.370 & 2.753 & 1.394 & 1.971 & 1.092 & 3.467 & 1.473 & 3.106 & 1.381\\
ETTm1 & 24 & \best{0.217} & \best{0.299} & \secondbest{0.232} & \secondbest{0.314} & 0.453 & 0.444 & 0.522 & 0.472 & 0.246 & 0.329 & 0.323 & 0.369 & 0.522 & 0.472\\
 & 48 & \best{0.279} & \best{0.339} & \secondbest{0.311} & \secondbest{0.368} & 0.592 & 0.521 & 0.695 & 0.567 & 0.381 & 0.386 & 0.494 & 0.503 & 0.542 & 0.508\\
 & 96 & \best{0.302} & \best{0.357} & \secondbest{0.360} & \secondbest{0.402} & 0.635 & 0.554 & 0.731 & 0.595 & 0.378 & 0.419 & 0.678 & 0.614 & 0.666 & 0.578\\
 & 288 & \best{0.377} & \best{0.398} & \secondbest{0.450} & \secondbest{0.467} & 0.693 & 0.597 & 0.818 & 0.649 & 0.472 & 0.486 & 1.056 & 0.786 & 0.991 & 0.735\\
 & 672 & \best{0.429} & \best{0.424} & \secondbest{0.612} & \secondbest{0.563} & 0.782 & 0.653 & 0.932 & 0.712 & 0.620 & 0.574 & 1.192 & 0.926 & 1.032 & 0.756\\
ETTm2 & 24 & \best{0.104} & \best{0.205} & \secondbest{0.108} & \secondbest{0.223} & 0.180 & 0.293 & 0.185 & 0.297 & 0.122 & 0.244 & 0.173 & 0.301 & 0.180 & 0.324\\
 & 48 & \best{0.138} & \best{0.236} & \secondbest{0.164} & \secondbest{0.285} & 0.244 & 0.350 & 0.264 & 0.360 & 0.183 & 0.305 & 0.303 & 0.409 & 0.204 & 0.327\\
 & 96 & \best{0.174} & \best{0.265} & \secondbest{0.271} & \secondbest{0.376} & 0.360 & 0.427 & 0.389 & 0.458 & 0.294 & 0.394 & 0.365 & 0.453 & 3.041 & 1.330\\
 & 288 & \best{0.270} & \best{0.326} & \secondbest{0.716} & 0.646 & 0.723 & \secondbest{0.639} & 0.920 & 0.788 & 0.723 & 0.652 & 1.047 & 0.804 & 3.162 & 1.337\\
 & 672 & \best{0.354} & \best{0.381} & \secondbest{1.600} & \secondbest{0.979} & 1.753 & 1.007 & 2.164 & 1.135 & 1.899 & 1.073 & 3.126 & 1.302 & 3.624 & 1.484\\
Exchange & 24 & \best{0.026} & \best{0.110} & \secondbest{0.059} & \secondbest{0.172} & 0.108 & 0.252 & 0.105 & 0.236 & 0.136 & 0.291 & 0.611 & 0.626 & 2.483 & 1.327\\
 & 48 & \best{0.042} & \best{0.143} & \secondbest{0.135} & \secondbest{0.265} & 0.200 & 0.341 & 0.162 & 0.270 & 0.250 & 0.387 & 0.680 & 0.644 & 2.328 & 1.256\\
 & 168 & \best{0.146} & \best{0.279} & 0.713 & 0.635 & 0.412 & 0.492 & \secondbest{0.397} & \secondbest{0.480} & 0.924 & 0.762 & 1.097 & 0.825 & 2.372 & 1.279\\
 & 336 & \best{0.340} & \best{0.422} & 1.409 & 0.938 & 1.339 & 0.901 & \secondbest{1.008} & \secondbest{0.866} & 1.774 & 1.063 & 1.672 & 1.036 & 3.113 & 1.459\\
 & 720 & \best{0.679} & \best{0.620} & \secondbest{1.628} & \secondbest{1.056} & 2.114 & 1.125 & 1.989 & 1.063 & 2.160 & 1.209 & 2.478 & 1.310 & 3.150 & 1.458\\
Weather & 24 & \best{0.101} & \best{0.145} & \secondbest{0.298} & \secondbest{0.359} & 0.308 & 0.364 & 0.320 & 0.373 & \secondbest{0.298} & 0.360 & 0.335 & 0.381 & 0.321 & 0.367\\
 & 48 & \best{0.128} & \best{0.181} & \secondbest{0.359} & \secondbest{0.410} & 0.375 & 0.417 & 0.380 & 0.421 & \secondbest{0.359} & 0.411 & 0.395 & 0.459 & 0.386 & 0.423\\
 & 168 & \best{0.194} & \best{0.244} & \secondbest{0.426} & \secondbest{0.461} & 0.496 & 0.506 & 0.479 & 0.495 & 0.464 & 0.491 & 0.608 & 0.567 & 0.491 & 0.501\\
 & 336 & \best{0.249} & \best{0.285} & 0.504 & 0.520 & 0.532 & 0.533 & 0.505 & 0.514 & \secondbest{0.497} & 0.517 & 0.702 & 0.620 & 0.502 & \secondbest{0.507}\\
 & 720 & \best{0.323} & \best{0.341} & 0.535 & 0.542 & 0.567 & 0.558 & 0.543 & 0.547 & 0.533 & 0.542 & 0.831 & 0.731 & \secondbest{0.498} & \secondbest{0.508}
\end{tblr}
\end{table*}
\begin{table*}[ht]
\centering
\caption{\redtext{\textbf{Linear evaluation on univariate time-series forecasting.} We use prediction lengths \(T \in \{24, 48, 168, 336, 720\}\) for ETTh1, ETTh2, Exchange, and Weather; and \(T \in \{24, 48, 96, 228, 672\}\) for ETTm1 and ETTm2. The best results are in \best{bold}, while the second-best are \secondbest{underlined}.}}
\label{tab:linear_eval_forecasting_univariate}
\begin{tblr}{
  rowsep = 1.1pt,
  colsep = 5pt,
  cells = {c},
  cells={font=\small},
  cell{1}{1} = {c=2,r=2}{},
  cell{1}{3} = {c=10}{},
  cell{1}{13} = {c=4}{},
  cell{2}{3} = {c=2}{},
  cell{2}{5} = {c=2}{},
  cell{2}{7} = {c=2}{},
  cell{2}{9} = {c=2}{},
  cell{2}{11} = {c=2}{},
  cell{2}{13} = {c=2}{},
  cell{2}{15} = {c=2}{},
  cell{3}{1} = {c=2}{},
  cell{34}{1} = {c=2}{},
  cell{4}{1} = {r=5}{},
  cell{9}{1} = {r=5}{},
  cell{14}{1} = {r=5}{},
  cell{19}{1} = {r=5}{},
  cell{24}{1} = {r=5}{},
  cell{29}{1} = {r=5}{},
  vline{2-3,5,7,9,11,13,15} = {1-39}{},
  hline{1-4,9,14,19,24,29,34} = {-}{},
}
Methods &  & Unsupervised
  Representation Learning &  &  &  &  &  &  &  &  &  & End-to-end
  Forecasting &  &  & \\
 &  & TimeDRL &  & SimTS &  & TS2Vec &  & TNC &  & CoST &  & Informer &  & TCN & \\
Metric &  & MSE & MAE & MSE & MAE & MSE & MAE & MSE & MAE & MSE & MAE & MSE & MAE & MSE & MAE\\
ETTh1 & 24 & \best{0.027} & \best{0.126} & \secondbest{0.036} & \secondbest{0.143} & 0.039 & 0.151 & 0.057 & 0.184 & 0.04 & 0.152 & 0.098 & 0.147 & 0.104 & 0.254\\
 & 48 & \best{0.040} & \best{0.152} & \secondbest{0.054} & \secondbest{0.176} & 0.062 & 0.189 & 0.094 & 0.239 & 0.060 & 0.186 & 0.158 & 0.319 & 0.206 & 0.366\\
 & 168 & \best{0.068} & \best{0.200} & \secondbest{0.084} & \secondbest{0.216} & 0.142 & 0.291 & 0.171 & 0.329 & 0.097 & 0.236 & 0.183 & 0.346 & 0.462 & 0.586\\
 & 336 & \best{0.084} & \best{0.228} & \secondbest{0.100} & \secondbest{0.239} & 0.160 & 0.316 & 0.179 & 0.345 & 0.112 & 0.258 & 0.222 & 0.387 & 0.422 & 0.564\\
 & 720 & \best{0.086} & \best{0.231} & \secondbest{0.126} & \secondbest{0.277} & 0.179 & 0.345 & 0.235 & 0.408 & 0.148 & 0.306 & 0.269 & 0.435 & 0.438 & 0.578\\
ETTh2 & 24 & \best{0.070} & \best{0.205} & \secondbest{0.077} & \secondbest{0.206} & 0.097 & 0.230 & 0.097 & 0.238 & 0.079 & 0.207 & 0.093 & 0.240 & 0.109 & 0.251\\
 & 48 & \best{0.097} & \best{0.241} & \secondbest{0.116} & \secondbest{0.259} & 0.124 & 0.274 & 0.131 & 0.281 & 0.118 & \secondbest{0.259} & 0.155 & 0.314 & 0.147 & 0.302\\
 & 168 & \best{0.166} & \best{0.323} & 0.191 & 0.340 & 0.198 & 0.355 & 0.197 & 0.354 & \secondbest{0.189} & \secondbest{0.339} & 0.232 & 0.389 & 0.209 & 0.366\\
 & 336 & \best{0.177} & \best{0.340} & \secondbest{0.199} & \secondbest{0.354} & 0.205 & 0.364 & 0.207 & 0.366 & 0.206 & 0.360 & 0.263 & 0.417 & 0.237 & 0.391\\
 & 720 & 0.222 & 0.378 & 0.212 & \secondbest{0.370} & 0.208 & 0.371 & \secondbest{0.207} & \secondbest{0.370} & 0.214 & 0.371 & 0.277 & 0.431 & \best{0.200} & \best{0.367}\\
ETTm1 & 24 & \best{0.012} & \best{0.080} & \secondbest{0.013} & \secondbest{0.084} & 0.016 & 0.093 & 0.019 & 0.103 & 0.015 & 0.088 & 0.030 & 0.137 & 0.027 & 0.127\\
 & 48 & \best{0.019} & \best{0.105} & \secondbest{0.024} & \secondbest{0.112} & 0.028 & 0.126 & 0.045 & 0.162 & 0.025 & 0.117 & 0.069 & 0.203 & 0.040 & 0.154\\
 & 96 & \best{0.028} & \best{0.129} & 0.041 & \secondbest{0.143} & 0.045 & 0.162 & 0.054 & 0.178 & \secondbest{0.038} & 0.147 & 0.194 & 0.372 & 0.097 & 0.246\\
 & 288 & \best{0.051} & \best{0.173} & 0.098 & \secondbest{0.207} & 0.095 & 0.235 & 0.142 & 0.290 & \secondbest{0.077} & 0.209 & 0.401 & 0.544 & 0.305 & 0.455\\
 & 672 & \best{0.070} & \best{0.201} & 0.117 & \secondbest{0.242} & 0.142 & 0.290 & 0.136 & 0.290 & \secondbest{0.113} & 0.257 & 0.277 & 0.431 & 0.200 & 0.367\\
ETTm2 & 24 & \secondbest{0.024} & \secondbest{0.104} & \best{0.022} & \best{0.099} & 0.038 & 0.139 & 0.045 & 0.151 & 0.027 & 0.112 & 0.036 & 0.141 & 0.048 & 0.153\\
 & 48 & \secondbest{0.048} & \secondbest{0.155} & \best{0.045} & \best{0.149} & 0.069 & 0.194 & 0.080 & 0.201 & 0.054 & 0.159 & 0.069 & 0.200 & 0.063 & 0.191\\
 & 96 & \best{0.065} & \best{0.187} & \secondbest{0.068} & \secondbest{0.189} & 0.089 & 0.225 & 0.094 & 0.229 & 0.072 & 0.196 & 0.095 & 0.240 & 0.129 & 0.265\\
 & 288 & \best{0.117} & \best{0.258} & 0.160 & \secondbest{0.272} & 0.161 & 0.306 & 0.155 & 0.309 & \secondbest{0.153} & 0.307 & 0.211 & 0.367 & 0.208 & 0.352\\
 & 672 & \best{0.172} & \best{0.322} & 0.249 & 0.334 & 0.201 & 0.351 & 0.197 & 0.352 & \secondbest{0.183} & \secondbest{0.329} & 0.267 & 0.417 & 0.222 & 0.377\\
Exchange & 24 & \best{0.026} & \best{0.122} & \secondbest{0.027} & \secondbest{0.128} & 0.033 & 0.142 & 0.082 & 0.227 & 0.028 & 0.128 & 0.103 & 0.262 & 0.033 & 0.139\\
 & 48 & \best{0.047} & \best{0.163} & 0.049 & \secondbest{0.169} & 0.059 & 0.191 & 0.116 & 0.268 & \secondbest{0.048} & \secondbest{0.169} & 0.121 & 0.283 & 0.060 & 0.188\\
 & 168 & 0.174 & \best{0.309} & \best{0.158} & \secondbest{0.314} & 0.180 & 0.340 & 0.275 & 0.411 & \secondbest{0.161} & 0.319 & 0.168 & 0.337 & 0.214 & 0.350\\
 & 336 & 0.412 & \best{0.485} & \best{0.382} & \secondbest{0.488} & 0.465 & 0.533 & 0.579 & 0.582 & \secondbest{0.399} & 0.497 & 1.672 & 1.036 & 0.476 & 0.527\\
 & 720 & \best{1.070} & \best{0.793} & 1.600 & 1.016 & 1.357 & 0.931 & 1.570 & 1.024 & 1.639 & 1.044 & 2.478 & 1.310 & \secondbest{1.166} & \secondbest{0.830}\\
Weather & 24 & \best{0.005} & \best{0.052} & 0.098 & 0.214 & \secondbest{0.096} & 0.215 & 0.102 & 0.221 & \secondbest{0.096} & \secondbest{0.213} & 0.117 & 0.251 & 0.109 & 0.217\\
 & 48 & \best{0.002} & \best{0.032} & \secondbest{0.136} & \secondbest{0.260} & 0.140 & 0.264 & 0.139 & 0.264 & 0.138 & 0.262 & 0.178 & 0.318 & 0.143 & 0.269\\
 & 168 & \best{0.017} & \best{0.095} & \secondbest{0.120} & 0.328 & 0.207 & 0.335 & 0.198 & 0.328 & 0.207 & 0.334 & 0.266 & 0.398 & 0.188 & \secondbest{0.319}\\
 & 336 & \best{0.067} & \best{0.190} & 0.221 & 0.349 & 0.231 & 0.360 & 0.215 & 0.347 & 0.230 & 0.356 & 0.197 & 0.416 & \secondbest{0.192} & \secondbest{0.320}\\
 & 720 & \best{0.126} & \best{0.268} & 0.235 & 0.365 & 0.233 & 0.365 & 0.219 & 0.353 & 0.242 & 0.370 & 0.359 & 0.466 & \secondbest{0.198} & \secondbest{0.329}
\end{tblr}
\end{table*}

\paragraph{Evaluation Metrics for Time-Series Classification}
In time-series classification, we utilize accuracy (ACC), macro-averaged F1-score (MF1), and Cohen's Kappa coefficient (\(\kappa\)) as evaluation metrics across all experiments.
Accuracy is defined as
\begin{equation}
\text{ACC} = \frac{\text{TP} + \text{TN}}{\text{TP} + \text{TN} + \text{FP} + \text{FN}},
\end{equation}where TP, TN, FP, and FN represent true positive, true negative, false positive, and false negative, respectively.
The macro-averaged F1-score is calculated as:
\begin{equation}
\text{MF1} = \frac{2 \times \text{P} \times \text{R}}{\text{P} + \text{R}},
\end{equation}
where Precision (P) is 
\begin{equation}
\text{P} = \frac{\text{TP}}{\text{TP} + \text{FP}},
\end{equation}
and Recall (R) is
\begin{equation}
\text{R} = \frac{\text{TP}}{\text{TP} + \text{FN}}.
\end{equation}
Cohen’s Kappa coefficient is determined by:
\begin{equation}
\kappa = \frac{\text{ACC} - p_e}{1 - p_e},
\end{equation}
where \(p_e\) is the hypothetical probability of chance agreement, calculated using
\begin{equation}
p_e = \frac{(\text{TP}+\text{FN}) \times (\text{TP}+\text{FP}) + (\text{FP}+\text{TN}) \times (\text{FN}+\text{TN})}{N^2},
\end{equation}
with \(N\) denoting the total number of samples.
\redtext{
The coefficient ranges from \(-1\) (total disagreement) to \(1\) (perfect agreement), with \(0\) indicating no agreement beyond chance.
Cohen's Kappa (\(\kappa\)) is crucial for evaluating classifiers on imbalanced datasets because it adjusts for chance agreement.
This reveals when a classifier's performance is akin to random chance (\(\kappa\) close to \(0\)) or when \(\kappa\) is negative, indicating performance worse than random, providing critical insights into handling class imbalance.
}

\begin{table*}
\centering
\caption{\redtext{\textbf{Linear evaluation on time-series classification.} The best results are in \best{bold}, while the second-best are \secondbest{underlined}.}}
\label{tab:linear_eval_classification}
\begin{tblr}{
  rowsep = \myrowsep,
  cells={font=\small},
  cells = {c},
  cell{2}{1} = {r=3}{},
  cell{5}{1} = {r=3}{},
  cell{8}{1} = {r=3}{},
  cell{11}{1} = {r=3}{},
  cell{14}{1} = {r=3}{},
  cell{17}{1} = {r=3}{},
  vline{2-10} = {-}{},
  hline{1,2,5,8,11,14,17} = {-}{},
}
Dataset & Metric & TimeDRL & MHCCL & CCL & SimCLR & BYOL & TS2Vec & TSTCC & T-Loss\\
FingerMovements & ACC & \best{64.00} & \secondbest{52.09} & 50.23 & 49.20 & 49.60 & 50.00 & 50.17 & 50.50\\
 & MF1 & \best{63.77} & \secondbest{50.51} & 47.32 & 43.08 & 49.38 & 49.99 & 49.07 & 50.33\\
 & \(\kappa\) & \best{28.26} & \secondbest{17.87} & 9.18 & -0.69 & -0.49 & -0.01 & 0.33 & 4.01\\
PenDigits & ACC & \secondbest{98.00} & \best{98.69} & 91.27 & 89.24 & 94.93 & 97.83 & 97.44 & 97.86\\
 & MF1 & \secondbest{98.01} & \best{98.71} & 88.61 & 89.17 & 94.96 & 97.80 & 97.45 & 97.87\\
 & \(\kappa\) & \best{97.78} & 97.43 & 87.66 & 88.04 & 94.37 & 97.59 & 97.16 & \secondbest{97.63}\\
HAR & ACC & 89.01 & \best{91.60} & 86.84 & 81.06 & 89.46 & 90.47 & 89.22 & \secondbest{91.06}\\
 & MF1 & 89.41 & \best{91.77} & 83.56 & 80.62 & 89.31 & 90.46 & 89.23 & \secondbest{90.94}\\
 & \(\kappa\) & 86.79 & \best{89.90} & 81.46 & 77.25 & 87.33 & 89.15 & 87.03 & \secondbest{89.26}\\
Epilepsy & ACC & 97.78 & \secondbest{97.85} & 95.47 & 93.00 & \best{98.08} & 96.32 & 97.19 & 96.94\\
 & MF1 & \secondbest{96.53} & 95.44 & 91.38 & 88.09 & \best{96.99} & 94.27 & 95.47 & 95.20\\
 & \(\kappa\) & \secondbest{93.07} & 91.08 & 79.42 & 76.27 & \best{93.99} & 88.54 & 90.94 & 90.41\\
WISDM & ACC & 91.45 & \best{93.60} & 85.18 & 83.04 & 87.84 & \secondbest{92.33} & 81.48 & 91.48\\
 & MF1 & 82.37 & \best{91.70} & 81.22 & 75.83 & 84.02 & \secondbest{90.27} & 69.17 & 88.79\\
 & \(\kappa\) & 87.85 & \best{90.96} & 79.19 & 75.15 & 82.43 & \secondbest{90.36} & 73.13 & 87.79
\end{tblr}
\end{table*}

\subsubsection{\textbf{Baselines}}
\paragraph{Baselines for Time-Series Forecasting}
\textbf{SimTS} \cite{simts} simplifies time-series forecasting by learning to predict future outcomes from past data in a latent space, without depending on negative pairs or specific assumptions about the time-series characteristics.
\textbf{TS2Vec} \cite{ts2vec} is the first universal framework for time-series representation learning, focusing on differentiating multi-scale contextual information at both instance and timestamp levels, proving effective across a range of time-series tasks.
\textbf{TNC} \cite{tnc} employs the Augmented Dickey-Fuller test for identifying temporal neighborhoods, and adopts Positive-Unlabeled learning to mitigate sampling bias.
\textbf{CoST} \cite{cost} integrates contrastive losses from both time and frequency domains, enabling the learning of distinct trend and seasonal representations respectively.

Alongside unsupervised representation learning methods, we also incorporate two end-to-end learning approaches.
In these models, both the representation learning and forecasting components are integrated and trained simultaneously in an end-to-end manner.
\textbf{Informer} \cite{informer} introduces ProbSparse self-attention and distilling operations, addressing the quadratic time complexity and memory usage challenges in the standard Transformer.
\textbf{TCN} \cite{tcn} merges dilations and residual connections with causal convolutions, essential for autoregressive prediction.

\paragraph{Baselines for Time-Series Classification}
\textbf{MHCCL} \cite{mhccl} utilizes semantic data from a hierarchical structure in multivariate time-series, using hierarchical clustering to improve positive and negative sample pairing.
\textbf{CCL} \cite{ccl} adopts a clustering-based approach for representation learning, leveraging labels derived from clustering and constraints to develop discriminative features.
\textbf{SimCLR} \cite{simclr} employs contrastive learning, treating augmented views of the same instance as positive pairs and different instances in a minibatch as negatives.
\textbf{BYOL} \cite{byol} uses two interacting networks, online and target, to learn from augmented views of an image, with the online network predicting the target's representation under varied augmentations and the target network being a slow-moving average of the online network.
\textbf{TS2Vec} \cite{ts2vec}, recognized as a universal framework for time-series representation learning, is also included in our analysis for time-series classification.
\textbf{TS-TCC} \cite{ts_tcc} generates two views through strong and weak augmentations, then learns representations by contrasting these views temporally and contextually.
\textbf{T-Loss} \cite{t_loss} trains representations through triplet loss with time-based negative sampling.

\subsubsection{\textbf{Implementation Details}}
We partition the dataset into three segments: 60\% for training, 20\% for validation, and 20\% for testing, except when a predefined train-test split exists.
We employ AdamW \cite{adamw} optimizer with weight decay.
The experiments are conducted on an NVIDIA GeForce RTX 3070 GPU.
We use Transformer encoder as the encoder's architecture \(f_\theta\).
We design the timestamp-predictive head \(p_\theta\) using a linear layer, and we design the instance-contrastive head \(p_\theta\) using a two-layer bottleneck MLP with BatchNorm and ReLU in the middle.
In the time-series forecasting task, we incorporate channel-independence alongside patching, a concept also introduced by PatchTST \cite{patchtst}.
This method treats multivariate time-series as multiple univariate series, processed collectively by a single model. Although channel-mixing models directly exploit cross-channel data, channel-independence captures cross-channel interactions indirectly through shared weights.
We observed that channel-independence significantly enhances performance in time-series forecasting, leading to its integration in our experiments.
However, for time-series classification, we found that omitting channel-independence yielded better results.

\subsection{Linear Evaluation on Time-Series Forecasting}
\label{sec:linear_eval_forecasting}
To assess the effectiveness of TimeDRL's timestamp-level embeddings, we conduct a linear evaluation on time-series forecasting.
This involves pre-training the encoder using pretext tasks, then freezing the encoder weights and attaching a linear layer for training on the downstream forecasting task.
Following the experimental settings used in SimTS \cite{simts}, we set various prediction lengths \(T \in \{24, 48, 168, 336, 720\}\) for datasets like ETTh1, ETTh2, Exchange, and Weather; and \(T \in \{24, 48, 96, 228, 672\}\) for ETTm1 and ETTm2.
The performance of TimeDRL in multivariate forecasting is summarized in Table \ref{tab:linear_eval_forecasting_multivariate}, where it shows an average improvement of 58.02\% in MSE compared to state-of-the-art methods.

\redtext{
Notably, TimeDRL surpasses the performance of all baselines, including SimTS, despite its contrastive learning objective being specifically designed for forecasting tasks.
Although SimTS is effective at predicting the latent representation of future data from historical data, its performance is limited by not accounting for randomness in the data.
In contrast, TimeDRL incorporates randomness, allowing it to capture temporal dynamics more effectively and enhance prediction accuracy.
The omission of randomness in SimTS results in reduced generalizability, as it fails to account for the inherent variability and unpredictability in real-world scenarios.
This is particularly evident with longer prediction lengths.
For example, on the ETTh2 dataset with a prediction length of 720, TimeDRL achieves a 78.96\% improvement in MSE over SimTS, underscoring its significant advancement in long-term forecasting accuracy.
}
Additionally, to assess the model's capability with univariate time-series data, we conducted experiments on univariate forecasting.
The results, as shown in Table \ref{tab:linear_eval_forecasting_univariate}, reveal an average improvement of 29.09\% in MSE, further validating the versatility of TimeDRL.

\begin{figure}
    \centering
    \includegraphics[width=\columnwidth]{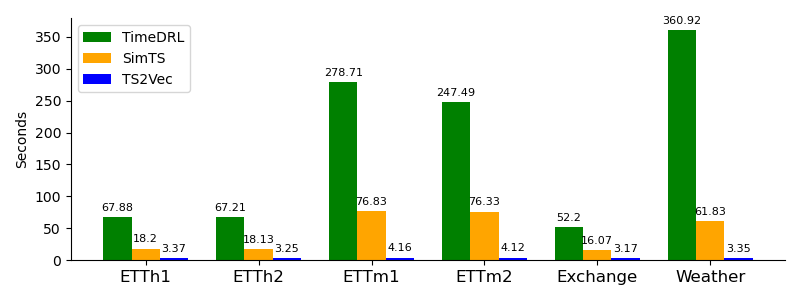}
    \caption{\redtext{\textbf{Comparison of training time (in seconds) in the pre-training stage on forecasting datasets.}}}
    \label{fig:running_time}
\end{figure}

\redtext{
The execution time of TimeDRL, alongside the top-performing baselines SimTS and TS2Vec, is depicted in Fig. \ref{fig:running_time}, utilizing an NVIDIA GeForce RTX 3070 GPU for evaluation.
For a fair comparison across all methods, we set the batch size to 32, the number of epochs to 10, and the sequence length \(T\) to 512.
SimTS and TS2Vec use fast, convolutional-based encoders, while TimeDRL utilizes a Transformer encoder, known for its superior ability to capture temporal dependencies but with longer execution times.
To improve efficiency, TimeDRL integrates a patching mechanism, reducing input sequence length from \(L\) to \(\lfloor (L-P)/S \rfloor + 2\), thus lowering computational and memory demands quadratically.
Despite TimeDRL's longer execution time compared to convolutional-based counterparts, its patching mechanism significantly closes the efficiency gap, enhancing its ability to capture complex temporal dependencies without compromising performance.
}

\begin{figure}
    \centering
    \includegraphics[width=\columnwidth]{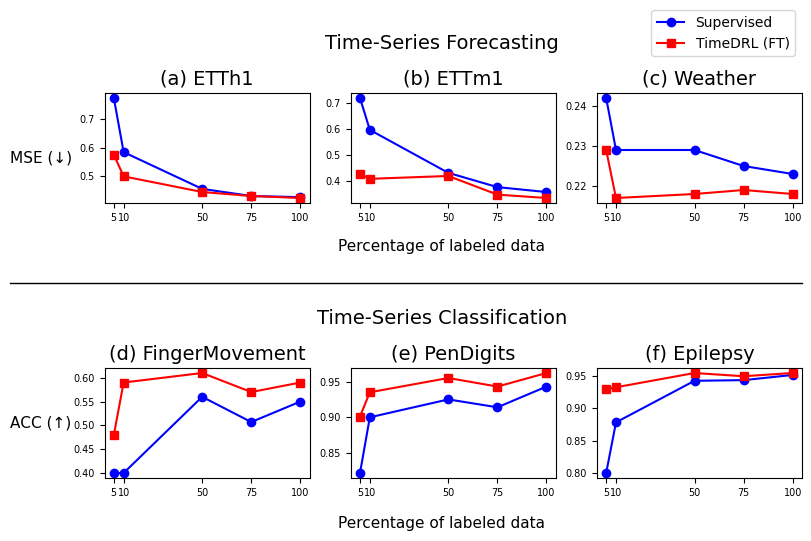}
    \caption{\redtext{\textbf{Semi-supervised learning.} To simulate real-world scenarios with limited availability of labeled data, we experiment with different portions of the labels in our datasets. The top sections (a-c) represent time-series forecasting, while the bottom sections (d-f) correspond to time-series classification. 'TimeDRL (FT)' indicates that we fine-tune the encoder during the downstream tasks.}}
    \label{fig:semi_supervised}
\end{figure}

\subsection{Linear Evaluation on Time-Series Classification}
\label{sec:linear_eval_classification}
To evaluate the efficacy of TimeDRL's instance-level embeddings, we utilize a linear evaluation approach for time-series classification.
Following a similar approach as in our forecasting evaluation, we first train the encoder using self-supervised learning, freeze its weights, and then attach a linear layer for training on classification tasks.
The performance of TimeDRL in time-series classification is detailed in Table \ref{tab:linear_eval_classification}, demonstrating an average accuracy improvement of 1.48\% over state-of-the-art methods.
For datasets where baseline models already achieve around 90\% accuracy or higher, TimeDRL consistently maintains comparable results.
Notably, on the challenging FingerMovements dataset, where baseline models generally underperform, TimeDRL shows a significant 22.86\% improvement in accuracy and a 58.13\% enhancement in Cohen's Kappa coefficient.
Furthermore, TimeDRL's capability with univariate data is evident in the Epilepsy dataset, which is only marginally lower than the best baseline method by 0.07\% in accuracy.

\subsection{Semi-supervised learning}
\label{sec:semi_supervised_learning}
The most practical application of self-supervised learning in real-world scenarios is semi-supervised learning, where labeled data are limited, but a vast amount of unlabeled data exists.
Traditional supervised learning methods focus solely on limited labeled data, neglecting the untapped potential of unlabeled data.
Self-supervised learning shines in this context by enabling the use of extensive unlabeled data for powerful representation learning.
We first train an encoder on a large unlabeled dataset to learn rich representations, followed by fine-tuning with limited labeled data alongside the downstream task head.
Unlike the linear evaluation in previous sections where the encoder weights are frozen, in this real-world application scenario, the encoder weights are adjusted during fine-tuning.
To emulate the situation of limited labeled data availability, we randomly withhold a portion of labels in our datasets.
The comparative results of using only labeled data (supervised learning) versus combining both unlabeled and labeled data (TimeDRL with fine-tuning) are illustrated in Fig. \ref{fig:semi_supervised}.
\redtext{
The results indicate that incorporating unlabeled data with TimeDRL significantly boosts performance in both forecasting (with MSE) and classification (with accuracy), particularly as the proportion of available labeled data decreases.
This trend highlights TimeDRL's efficiency in utilizing unlabeled data to boost performance as the availability of labeled data reduces.
}
Remarkably, the benefits of TimeDRL's pre-training stage are evident even when 100\% of the labels are available.

\subsection{Ablation Study}
\subsubsection{Pretext Tasks}
\begin{figure}[t]
    \centering
    \includegraphics[width=\columnwidth]{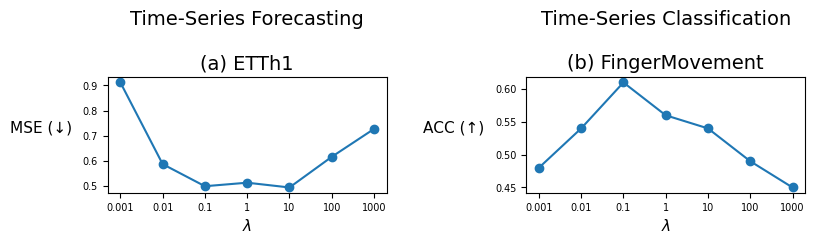}
    \caption{\redtext{\textbf{Sensitivity analysis on \(\lambda\).} When \(\lambda\) is smaller, the focus is more on predictive loss (\(\mathcal{L}_{P}\)); conversely, a larger \(\lambda\) shifts the focus towards contrastive loss (\(\mathcal{L}_{C}\)).}}
    \label{fig:lambda}
\end{figure}

In TimeDRL, we strategically employ two pretext tasks to optimize both timestamp-level and instance-level embeddings.
The timestamp-predictive task focuses on applying loss specifically to timestamp-level embeddings, while the instance-contrastive task targets instance-level embeddings.
\redtext{
Our experiment conducts a sensitivity analysis of the lambda parameter in Equation (\ref{eq:overall_loss}) to assess its impact on the effectiveness of representation learning.
According to the results in Fig. \ref{fig:lambda}, it is evident that combining both pretext tasks yields the best performance in both forecasting and classification tasks, highlighting the value of each task in enhancing dual-level embeddings. 
Notably, the instance-contrastive task, despite primarily optimizing instance-level embeddings, significantly improves performance in time-series forecasting tasks, which depend on timestamp-level embeddings.
In scenarios where the instance-contrastive task contributes minimally to the overall loss (\(\lambda = 0.001\)), the MSE experiences a substantial increase compared to scenarios where both losses are leveraged equally (\(\lambda = 1\)).
A similar effect is observed in the time-series classification task, where neglecting the timestamp-predictive task in favor of the instance-contrastive task (\(\lambda = 1000\)) leads to a significant decrease in accuracy.
}
These findings underscore the significance of both pretext tasks in the overall efficacy of TimeDRL across different time-series applications.

\subsubsection{Data Augmentations}
The foundational principle of TimeDRL is to avoid any data augmentation to prevent the introduction of inductive biases.
Accordingly, our design of both the timestamp-predictive task and the instance-contrastive task does not involve any augmentation methods.
In our experiment, we aim to demonstrate the potential drawbacks of ignoring the issue of inductive bias.
In Table \ref{tab:ablation_data_aug}, we experiment with 6 time-series-specific data augmentation methods \cite{data_aug_methods, ts2vec}.
\textbf{Jittering} simulates sensor noise through additive Gaussian noise.
\textbf{Scaling} adjusts the magnitude of data by multiplying it with a random scalar.
\textbf{Rotation} modifies the dataset by permuting the order of features and potentially flipping the sign of feature values.
\textbf{Permutation} slices the data into segments, then randomly permutes these segments to create a new time-series instance.
\textbf{Masking} randomly sets values in the time-series data to zero.
\textbf{Cropping} removes the left and right regions of a time-series instance and fills the gaps with zeros to maintain the same sequence length.

\redtext{
In the experiment detailed in Table \ref{tab:ablation_data_aug}, the application of any augmentation method led to a decline in performance, with an average increase in MSE of 27.77\% for the ETTh1 dataset and 57.37\% for the Exchange dataset.
The most significant deterioration was observed with Rotation augmentation, where MSE increased by 68.15\% for the ETTh1 dataset and 174.46\% for the Exchange dataset.
}
TS2Vec \cite{ts2vec} addresses the inductive bias issue but still employs Masking and Cropping augmentations.
Our results indicate that these two methods are relatively less harmful compared to others, yet they still lead to performance degradation.
This experiment supports our initial assumption that the complete avoidance of augmentation methods is crucial for eliminating inductive bias, thereby ensuring the optimal performance of TimeDRL.

\begin{table}[t]
\centering
\caption{\redtext{\textbf{Ablation study on data augmentation.} We use the prediction length \(T = 168\). The best results are in \best{bold}.}}
\label{tab:ablation_data_aug}
\begin{tblr}{
  rowsep = \myrowsep,
  cells={font=\small},
  cells = {c},
  vline{2,3} = {-}{},
  hline{1,2,9} = {-}{},
}
Data Augmentation & ETTh1 & Exchange\\
None (\textit{Ours}) & \best{0.418} & \best{0.146}\\
Jitter & 0.462 (+10.36\%) & 0.149 (+2.06\%)\\
Scaling & 0.534 (+27.67\%) & 0.256 (+74.77\%)\\
Rotation & 0.703 (+68.15\%) & 0.402 (+174.46\%)\\
Permutation & 0.607 (+45.09\%) & 0.199 (+35.73\%)\\
Masking & 0.438 (+4.76\%) & 0.160 (+9.47\%)\\
Cropping & 0.462 (+10.57\%) & 0.216 (+47.70\%)
\end{tblr}
\end{table}
\begin{table}[t]
\centering
\caption{\redtext{\textbf{Ablation study on pooling methods.} The best results are in \best{bold}.}}
\label{tab:ablation_pooling}
\begin{tblr}{
  rowsep = \myrowsep,
  colsep = 7.5pt,
  cells={font=\small},
  cells = {c},
  vline{2,3} = {-}{},
  hline{1,2,6} = {-}{},
}
Pooling Method & FingerMovements & Epilepsy\\
{[}CLS] (\textit{Ours}) & \best{63.00} & \best{95.83}\\
Last & 57.00 (-9.52\%) & 79.78  (-16.75\%)\\
GAP & 51.00 (-19.05\%) & 79.78  (-16.75\%)\\
All & 60.00 (-4.76\%) & 79.78  (-16.75\%)
\end{tblr}
\end{table}

\subsubsection{Pooling Methods}
In TimeDRL, we employ a dedicated [CLS] token strategy to derive instance-level embeddings directly from patched time-series data.
Despite this approach, we recognize the theoretical possibility of extracting instance-level embeddings from timestamp-level embeddings using pooling methods.
To explore this, we conducted experiments with 3 other different pooling strategies for instance-level embeddings, as detailed in Table \ref{tab:ablation_data_aug}.
\textbf{Last} utilizes the last timestamp-level embedding as the instance-level representation.
\textbf{GAP} employs global average pooling, averaging the timestamp-level embeddings across the time axis to aggregate the instance-level embedding.
\textbf{All} flattens all timestamp-level embeddings to create a singular instance-level representation.

\redtext{
The results in Table \ref{tab:ablation_pooling} reveal that employing strategies other than the [CLS] token strategy used by TimeDRL results in an average decrease in accuracy of 11.11\% for the FingerMovements dataset and 16.75\% for the Epilepsy dataset, highlighting the superior performance of the [CLS] token approach.
}
The least effective pooling method is GAP, commonly used in the time-series domain \cite{ts2vec}.
This method experiences the most significant performance drop due to the anisotropy problem.
These results demonstrate the importance of disentangling timestamp-level and instance-level embeddings, a key factor contributing to TimeDRL's superior performance compared to other baseline methods.

\begin{table}[t]
\centering
\caption{\redtext{\textbf{Ablation study on the architecture of the backbone encoder.} We use the prediction length \(T = 168\). The best results are in \best{bold}.}}
\label{tab:ablation_backbone}
\begin{tblr}{
  rowsep = \myrowsep,
  colsep = 3pt,
  cells={font=\small},
  cells = {c},
  vline{2,3} = {-}{},
  hline{1,2,8} = {-}{},
}
Backbones & ETTh1 & Exchange \\
Transformer Encoder (\textit{Ours}) & \best{0.418} & \best{0.146}\\
Transformer Decoder & 0.465 (+11.26\%) & 0.159 (+8.28\%)\\
ResNet & 0.576 (+37.76\%) & 0.160 (+9.50\%)\\
TCN & 0.517 (+23.72\%) & 0.148 (+1.13\%)\\
LSTM & 0.451 (+7.84\%) & 0.160 (+9.46\%)\\
Bi-LSTM & 0.443 (+5.92\%) & 0.153 (+4.62\%)
\end{tblr}
\end{table}
\begin{table}[t]
\centering
\caption{\redtext{\textbf{Ablation study on the stop gradient operation.} The best results are in \best{bold}.}}
\label{tab:ablation_stop_gradient}
\begin{tblr}{
  rowsep = \myrowsep,
  colsep = 7.5pt,
  cells={font=\small},
  cells = {c},
  vline{2,3} = {-}{},
  hline{1,2,4} = {-}{},
}
Stop Gradient & FingerMovements & Epilepsy\\
w/ SG (\textit{Ours}) & \best{63.00} & \best{95.83}\\
w/o SG & 56.00 (-11.11\%) & 79.78  (-16.75\%)
\end{tblr}
\end{table}

\subsubsection{Encoder Architectures}
Transformers are well-known for their success in downstream time-series tasks \cite{fedformer, patchtst}, but for self-supervised learning in time-series, CNN-based \cite{ts_tcc, ts2vec} and RNN-based \cite{tnc} models are usually chosen over Transformers.
TimeDRL is designed to make the best use of Transformers in self-supervised learning for time-series data, aiming to show the Transformer's strong capabilities in this field.
Within TimeDRL, the Transformer encoder is utilized as the core encoder.
To benchmark its performance against other encoder architectures, we conducted experiments with 5 different models, detailed in Table \ref{tab:ablation_backbone}.
\textbf{Transformer Decoder} employs a similar architecture to the Transformer encoder, with the key distinction of using masked self-attention. This ensures that each timestamp's embedding attends only to preceding timestamps, not subsequent ones.
\textbf{ResNet} adopts the well-known architecture from computer vision's ResNet18, modifying it with one-dimensional convolutions suitable for time-series data.
\textbf{TCN} \cite{tcn} combines dilations and residual connections with causal convolutions, specifically tailored for autoregressive prediction in time-series.
\textbf{LSTM} uses Long Short-Term Memory units for capturing dependencies in sequential data. A uni-directional LSTM is utilized, focusing on past and present data to prevent future data leakage.
\textbf{Bi-LSTM} follows the LSTM structure but incorporates bi-directional processing, allowing the model to integrate information from both past and future timestamps.

\redtext{
The findings presented in Table \ref{tab:ablation_backbone} indicate that not using the Transformer encoder, in combination with our two pretext tasks, results in worse model performance. 
Specifically, this results in an average increase of 17.30\% in MSE for the ETTh1 dataset and 6.60\% for the Exchange dataset compared to when the Transformer encoder is utilized, underscoring its efficacy over other architectures.
In contrast, compared to the Transformer encoder, the Transformer Decoder shows a decline in performance, with an 11.26\% increase in MSE for the ETTh1 dataset and 8.28\% for the Exchange dataset.
}
This difference highlights the critical role of bidirectional self-attention in achieving a comprehensive understanding of the entire sequence.
Similarly, when comparing LSTM with Bi-LSTM, the latter exhibits enhanced performance due to its ability to access both past and future information.
These findings highlight the importance of having full temporal access for each timestamp and confirm the Transformer encoder's strong capability of acquiring effective time-series representations.

\subsubsection{Stop Gradient}
To address sampling bias, our approach includes an extra prediction head coupled with a stop-gradient operation.
This asymmetric design, featuring one pathway with the additional prediction head and another with the stop-gradient, proves effective in preventing model collapse, as demonstrated in studies like \cite{simsiam, avoid_collapse}.
\redtext{
The results in Table \ref{tab:ablation_stop_gradient} show that eliminating the stop-gradient operation leads to a significant drop in accuracy of 11.11\% for the FingerMovements dataset and 16.75\% for the Epilepsy dataset, underscoring the crucial function of the stop-gradient element in this asymmetric architecture.
}

\section{Conclusion}
This paper introduces TimeDRL, a novel multivariate time-series representation learning framework with disentangled dual-level embeddings.
Our framework disentangles timestamp-level and instance-level embeddings, making it versatile for various time-series tasks such as forecasting and classification.
At its core, TimeDRL utilizes a [CLS] token strategy to extract contextualized instance-level embeddings from patched time-series data.
Two pretext tasks are introduced to optimize representations: the timestamp-predictive task focuses on optimizing timestamp-level embeddings with predictive loss, while the instance-contrastive task aims to optimize instance-level embeddings with contrastive loss.
To address the challenge of inductive bias, TimeDRL avoids the direct application of any data augmentations in both pretext tasks.
In the timestamp-predictive task, TimeDRL relies on the reconstruction error of patched time-series data, deliberately avoiding any input masking.
For the instance-contrastive task, TimeDRL exploits the randomness of dropout layers to create two distinct embedding views from a single data input.
Comprehensive experiments on 6 forecasting datasets and 5 classification datasets demonstrate TimeDRL's superior performance, delivering an average improvement of 58.02\% in MSE for forecasting and 1.48\% in accuracy for classification.
TimeDRL also supports the effective capability with limited labeled data in semi-supervised learning scenarios.
Moreover, extensive ablation studies with detailed analysis further validated the relative contributions of each component within TimeDRL's architecture.
\redtext{
In future work, we will enhance TimeDRL for time-series classification toward a more comprehensive foundation model.
Moreover, the comparisons with large language model (LLM)-based approaches will be investigated to study the impacts of LLMs in the time-series domain.
}


\bibliographystyle{IEEEtran}
\bibliography{reference}

\end{document}